\documentclass[pdflatex,sn-nature]{sn-jnl}%

\usepackage{graphicx}%
\usepackage{multirow}%
\usepackage{amsmath,amssymb,amsfonts}%
\usepackage{amsthm}%
\usepackage{mathrsfs}%
\usepackage[title]{appendix}%
\usepackage{xcolor}%
\usepackage{textcomp}%
\usepackage{manyfoot}%
\usepackage{booktabs}%
\usepackage{algorithm}%
\usepackage{algorithmicx}%
\usepackage{algpseudocode}%
\usepackage{listings}%

\theoremstyle{thmstyleone}%

\theoremstyle{thmstyletwo}%

\theoremstyle{thmstylethree}%

\usepackage[version=4]{mhchem} %
\usepackage{etoc}
\usepackage[resetlabels]{multibib}
\newcites{si}{Supplementary References}

\usepackage{makecell}

\begin{document}

\etocdepthtag.toc{mt}

\title[Article Title]{Substitution-Based Analysis of Structural Novelty for Generative Models of Materials}

\author[1]{\fnm{Masahiro} \sur{Negishi}}\email{m.negishi25@imperial.ac.uk}

\author*[1]{\fnm{Aron} \sur{Walsh}}\email{a.walsh@imperial.ac.uk}

\affil[1]{\orgdiv{Department of Materials}, \orgname{Imperial College London}, \orgaddress{\street{Exhibition Road}, \city{South Kensington}, \postcode{SW7 2AZ}, \state{London}, \country{United Kingdom}}}

\abstract{There has been rapid progress in generative artificial intelligence (AI) models for inorganic crystal design, which can efficiently generate large numbers of candidate compounds after being trained on databases of known crystals. However, it remains unclear whether they genuinely expand the accessible materials search space beyond conventional strategies such as elemental substitution within known structure types. We address this question by developing a workflow to assess whether AI-generated crystals are duplicates of training structures, reproducible by elemental substitution, or unmatched by either criterion. Applying this workflow to representative generative models reveals that 81--92\% of chemically valid and metastable generated crystals are either training duplicates or substitution-derived structures. This tendency is particularly strong in high-symmetry crystal systems, even though many possible structural prototypes remain unexplored. Further analysis of the underlying structural fingerprints shows that low-symmetry structures beyond duplication or substitution can be interpreted as interpolation in training-data-rich regions, while high-symmetry duplicates appear to result from memorisation in training-sparse regions. Our findings highlight a limitation in the current generation of models that exhibit a bias towards known structural prototypes in the high symmetry regions, but enable wider exploration of the low-symmetry structural space.}


\maketitle

\section{Introduction}

The discovery of inorganic crystals with useful properties has underpinned many of the technologies that shape modern society, as exemplified by semiconductors and lithium-ion batteries. 
Its importance is growing in light of urgent societal challenges such as climate change, where new materials are needed for applications including next-generation batteries and carbon capture \citep{rolnick2022tackling}.
Inorganic crystals synthesised to date are recorded in the Inorganic Crystal Structure Database (ICSD), which contains 245,944 entries as of version 2026.1 \citep{zagorac2019recent}. 
However, this number represents only a tiny fraction of the possible inorganic crystal space. 
For example, it is estimated that quaternary compounds formed from the first 103 elements generate over $10^{12}$ compositional combinations, or approximately $10^{10}$ after valency and electronegativity screening \citep{davies2016computational}. 
As this estimate excludes structural variation and higher-order compositions, the actual search space is far larger. 
Efficiently discovering inorganic crystals with desired properties from this nearly boundless space is therefore a fundamental goal of materials science. 
In this context, machine learning generative models have attracted increasing attention because they can learn from large databases of known crystals and rapidly generate numerous candidate compounds \citep{park2024has,cheng2025ai,de2025review,li2025materials,recatala2026generative}.

In principle, generative models are expected to explore a broader region of materials space than conventional approaches such as evolutionary algorithms (EAs) \citep{Tipton_2013, LONIE2011372, GLASS2006713, HAJINAZAR2026109910}. 
However, the structural novelty of (metastable) AI-generated crystals has been questioned, with some studies suggesting that their underlying structural motifs are often already known despite their novel compositions \citep{szymanski2025establishing,negishi2026continuoussunstableunique}. 
This raises an important question: \textbf{Do generative models genuinely expand the search space beyond what can be achieved by conventional strategies, such as elemental substitution within known structure types \citep{fischer2006predicting, hautier2011data, sciadv.abn4117, wang2021predicting, liu2024shotgun}?}
The most common approach to evaluating the structural novelty of generated crystals involves embedding the generated and reference compounds in continuous feature vectors and comparing them \citep{negishi2026continuoussunstableunique}. 
CrystalNN \citep{zimmermann2020local} and AMD \citep{widdowson2022average} vectors are frequently used for this purpose. 
However, these descriptors are not designed to directly address the research question considered here. 
Our primary interest is whether generative models can generate crystals that are difficult to reach using classical approaches. 
In other words, this study defines structural novelty by whether a generated crystal falls outside the search space readily reachable by conventional methods.

\begin{figure}
    \centering
    \includegraphics[width=\linewidth]{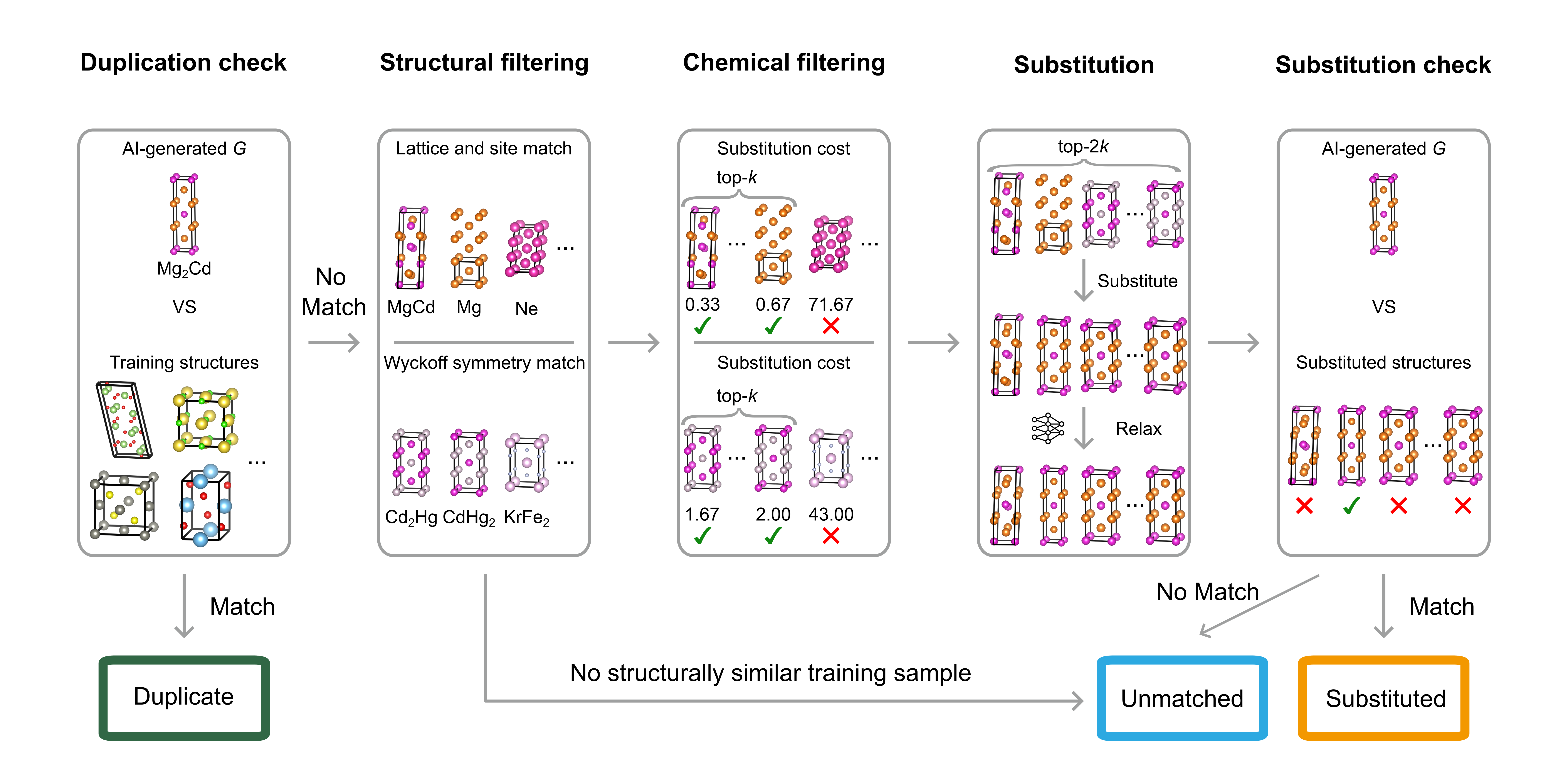}
    \caption{Workflow for classifying each AI-generated crystal as a Duplicate, Substituted, or Unmatched structure. Given an AI-generated crystal, $G$, we first compare it with all crystals in the training set to identify exact matches, which are classified as Duplicates. If no exact match is found, training crystals that are structurally similar to $G$ are identified based on lattice parameters and atomic coordinates or on Wyckoff symmetry. These candidates are then filtered according to their chemical similarity to $G$, defined as the average difference in the modified Pettifor chemical scale \citep{glawe2016optimal} between atoms at corresponding sites. Elemental substitution followed by a single structural relaxation using the MACE-MPA-0 interatomic potential \citep{batatia2025foundation} are then applied to the top-$2k$ selected training crystals, comprising the top-$k$ candidates from each of the lattice-and-site-matched set and the Wyckoff-symmetry-matched set. Finally, the substituted and relaxed structures are compared with $G$ to determine whether $G$ should be classified as Substituted or Unmatched. The MP20 dataset \citep{xie2022crystal} and the default value of $k=3$ are used in this study.}
    \label{fig:workflow}
\end{figure}

To address our research question directly, we develop a systematic workflow for examining whether AI-generated crystals can be reproduced from crystals in the training dataset using simple operations (Figure~\ref{fig:workflow}).
While EAs allow for diverse mutation and crossover operations to modify crystals \citep{cheng2025ai}, the present workflow is restricted to atomic substitution followed by a single final structural relaxation for simplicity.
For each AI-generated crystal, training crystals with high structural and chemical similarity are selected as initial structures.
The selected training crystals are then transformed by substituting their atoms with the elements of the AI-generated crystal and subsequently relaxed. 
The resulting structures are finally compared with the AI-generated crystal to determine whether it is accessible through substitution and relaxation alone.

Our experiments show that 81--92\% of chemically valid \citep{davies2016computational} and metastable crystals generated by current models are either duplicated from the training set or reproducible through elemental substitution. 
This tendency is particularly strong in high-symmetry crystal systems, even though a large fraction of potential high-symmetry structural prototypes remains unexplored.
Further analyses reveal that low-symmetry structures beyond duplication or substitution cluster in training-data-rich regions of structural space.
This suggests that models learn local structural environments from abundant training examples and use this knowledge to generate crystals with plausible local environments and complex global structures, effectively interpolating between training samples.
In training-sparse regions, however, the models appear more prone to memorisation, resulting in higher duplicate rates in high-symmetry systems. 
Overall, current crystal generative models show limited expansion beyond substitution-based search in high-symmetry regions, while offering higher potential for exploration in low-symmetry regions.

\section{Results}

\subsection{Classification Workflow}

\begin{table}
    \caption{Ten most frequent elemental substitutions observed between MatterGen-generated crystals \citep{MatterGen2025} classified as Substituted and their matched MP20 training structures. In our matching pipeline, the elements listed in the first row are replaced by those listed in the second row. The third row reports the substitution cost for each element pair, defined as the difference in the modified Pettifor chemical scale. This cost takes an integer value between 1 and 102, with smaller values denoting chemically more similar substitutions. All ten substitutions have low cost, indicating that frequent substitutions primarily occur between chemically similar elements.}
    \label{tab:top10_substitution}
    \centering
    \begin{tabular}{@{}lcccccccccc@{}}
        \toprule
        & 1 & 2 & 3 & 4 & 5 & 6 & 7 & 8 & 9 & 10 \\
        \midrule
        Element in the training structure & O & Si & Al & Fe & Ni & Ni & Au & F & Ga & Pd \\
        Element in the generated structure & F & Ge & Ga & Co & Cu & Co & Pd & O & Al & Pt \\
        Modified Pettifor scale difference & 5 & 1 & 1 & 1 & 1 & 1 & 1 & 5 & 1 & 1 \\
        \bottomrule
    \end{tabular}
\end{table}

This section outlines the workflow; full details are provided in Section~\ref{sec:method}. 
As illustrated in Figure~\ref{fig:workflow}, each AI-generated crystal $G$ is classified as Duplicate, Substituted, or Unmatched. 
We first identify Duplicates by comparing $G$ with all training structures using \texttt{StructureMatcher.fit} method in the \texttt{pymatgen} package \citep{ong2013python}. 
This step mirrors standard novelty evaluation for AI-generated samples, whereas the subsequent steps provide a more fine-grained assessment.
For non-Duplicate samples, we assess whether $G$ can be obtained from a training structure by elemental substitution followed by relaxation. 
Because optimal atomic-site mappings between a generated crystal $G$ and all training structures are not generally known in advance, and exhaustive comparison is computationally prohibitive, we first select training structures that are structurally similar to $G$ using two criteria: similarity in cell shape and fractional coordinates, and matching of space group and Wyckoff labels.
These candidates are then ranked by substitution cost, defined as the average modified Pettifor chemical scale difference \citep{glawe2016optimal} between matched atom pairs. 
The modified Pettifor chemical scale is a 1D ordering of elements that reflects chemical similarity: elements that are frequently interchangeable in the ICSD database are positioned close to one another. 
It is therefore well-suited for quantifying the chemical plausibility of elemental substitutions.
This screening reduces the number of post-substitution relaxations and prevents chemically implausible substitutions, which may be avoided in classical algorithms for searching new materials.
Table~\ref{tab:top10_substitution} lists the ten most frequent elemental substitutions between crystals generated by MatterGen \citep{MatterGen2025} and the MP20 training samples \citep{xie2022crystal}; all involve chemically similar element pairs, as intended.
After retaining the top-$k$ (default is 3) candidates with minimum substitution cost from each structural similarity criterion, we apply elemental substitution and relaxation with the MACE-MPA-0 interatomic potential \citep{batatia2025foundation}. The resulting structures are finally compared with $G$ using \texttt{StructureMatcher} to determine whether it is Substituted or Unmatched.

\subsection{Overall Classification Results}

\begin{table}
    \caption{Classification of crystals generated by each model as Duplicate, Substituted, or Unmatched. Duplicate crystals exactly match structures in the MP20 training set \citep{xie2022crystal}, whereas Substituted crystals are reproducible from training structures by elemental substitution followed by a single structural relaxation. All remaining crystals are classified as Unmatched. Candidate sets were obtained by applying stability ($E_\mathrm{hull} \le 0.1$ [eV/atom]) and chemical-validity filters \citep{davies2016computational} to 10,000 samples generated by each model. A large proportion of the generated crystals fall into the Duplicate or Substituted categories.}
    \label{tab:classification}
    \centering
    \begin{tabular}{@{}lcccc@{}}
        \toprule
        Model & Duplicate (\%) & Substituted (\%) & Unmatched (\%) \\
        \midrule
        MatterGen \citep{MatterGen2025} & 20.5 & 60.6 & 18.9 \\
        DiffCSP++ \citep{jiao2024space} & 28.7 & 63.0 & 8.3 \\
        WyckoffTransformer \citep{kazeev2025wyckoff} & 26.6 & 59.7 & 13.7 \\
        Crystalite \citep{veljkovic2026crystalite} & 29.6 & 60.5 & 9.9 \\
        Chemeleon2 \citep{park2025guiding} & 8.3 & 80.3 & 11.4 \\
        MP20 Test & 5.4 & 88.4 & 6.2 \\
        \bottomrule
    \end{tabular}
\end{table}

The workflow shown in Figure~\ref{fig:workflow} was applied to crystals sampled from representative generative models (MatterGen \citep{MatterGen2025}, DiffCSP++ \citep{jiao2024space}, WyckoffTransformer \citep{kazeev2025wyckoff}, Crystalite \citep{veljkovic2026crystalite}, and Chemeleon2 \citep{park2025guiding}) trained on the MP20 dataset, which consists of metastable crystals with up to 20 atoms in the primitive unit cell \citep{xie2022crystal}. 
From each model, 10,000 crystals were sampled unconditionally and filtered to remove compositions that failed the charge-neutrality or Pauling electronegativity tests \citep{davies2016computational}. 
The remaining structures were relaxed using the MACE-MPA-0 interatomic potential \citep{batatia2025foundation}, after which thermodynamically unstable samples were excluded using a threshold of 0.1 [eV/atom] above the convex hull constructed from the Materials Project database \citep{jain2013commentary}.

Table~\ref{tab:classification} summarises the classification results for the metastable relaxed samples, with the MP20 test set included for reference.
Overall, 81--92\% of the generated samples are either duplicates of training structures or reproducible through elemental substitution. 
This pattern holds across all models, including Crystalite, which reports the highest mSUN score (metastable, unique, and novel) on the LeMat-GenBench leaderboard \citep{betala2025lemat} at the time of writing (18/06/2026), and Chemeleon2, which is explicitly finetuned via reinforcement learning (RL) to explore non-Duplicate structures. 
Although RL appears to reduce the Duplicate rate, it does not improve the Unmatched rate. 
In a typical novelty-evaluation pipeline based on the \texttt{pymatgen}'s \texttt{StructureMatcher}, both Substituted and Unmatched structures would be classified as novel because neither exactly matches a training structure. 
Our classification thus reveals a more nuanced picture of generative models’ capacity for structural exploration, showing that the majority of samples conventionally classified as novel can in fact be reproduced through elemental substitution.

A noteworthy observation is the high Substituted rate for the MP20 test set. 
This result is consistent with the historical practice in inorganic compound discovery, where new compounds have often been identified by substituting atoms in known crystal structures \citep{fischer2006predicting, hautier2011data, sciadv.abn4117, HICKS2019S1, liu2024shotgun, wang2021predicting}.
It is also worth noting that 5.4\% of the test set samples were identified as duplicates of training samples. 
Although \texttt{StructureMatcher} may classify two similar but distinct crystals as matching when their structural differences fall below the specified thresholds, the duplicate rate remained high at 4.2\% even after all tolerance thresholds were halved relative to their default values.
While MP20 is currently one of the most widely used datasets for training and evaluating crystal generative models, nearly five years have passed since its release in 2021 \citep{xie2022crystal}. 
Dataset calibration or reconstruction is beyond the scope of the present study, and we therefore do not attempt to recalibrate MP20 here. 
Nevertheless, our results highlight the need for improved dataset curation and updated benchmarks for evaluating crystal generative models.

\subsection{Classification Trends across Crystal Systems}

\begin{figure}
    \centering
    \includegraphics[width=\linewidth]{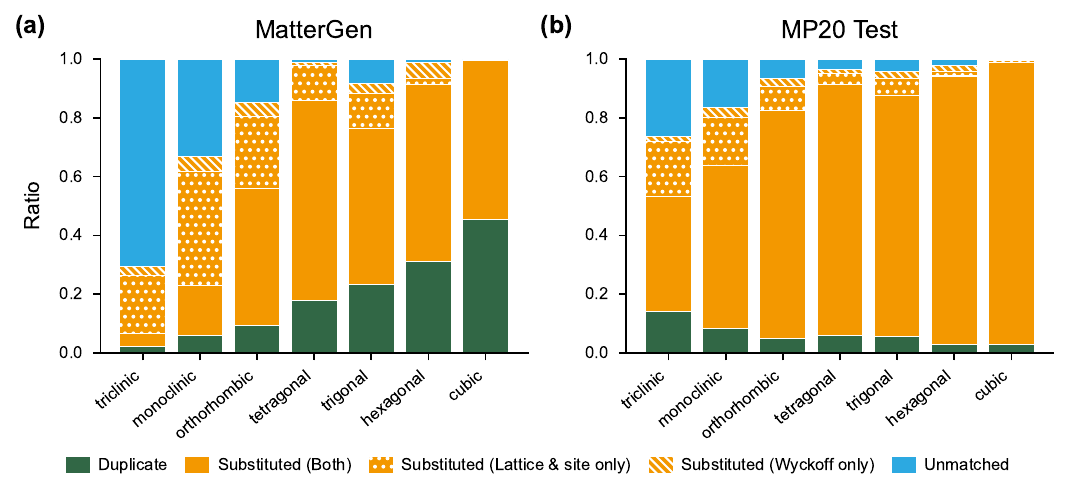}
    \caption{Classification of MatterGen-generated samples (\textbf{a}) and MP20 test samples (\textbf{b}) by crystal system. Parentheses in the legend denote the structural similarity criterion by which the training structures that ultimately matched each sample after elemental substitution and relaxation were identified. In both plots, the Unmatched ratio decreases with increasing crystal symmetry, with this trend being more pronounced for MatterGen. For MatterGen, Unmatched samples constitute only 1\% of the tetragonal and hexagonal systems and are absent in the cubic system. AI-generated structures also exhibit a substantially higher Duplicate ratio than the MP20 test samples in high-symmetry systems.}
    \label{fig:classification_by_crystal_system}
\end{figure}

\begin{figure}
    \centering
    \includegraphics[width=\linewidth]{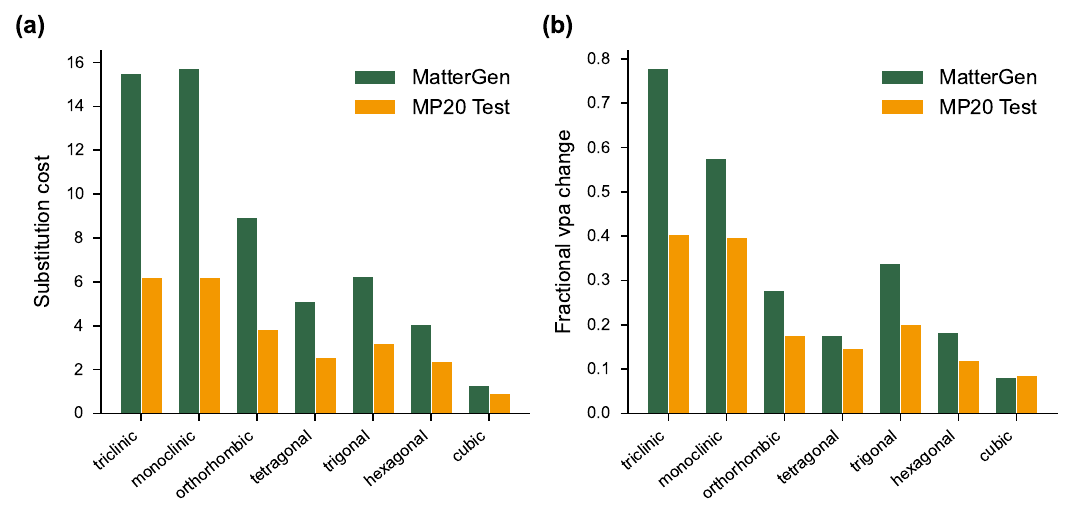}
    \caption{\textbf{a.} Average substitution cost between samples in each crystal system and their top-6 matched training structures. Lower substitution costs in high-symmetry systems indicate greater chemical similarity between substituted and original atoms. \textbf{b.} Average fractional change in volume per atom (VPA) during relaxation, defined as $\frac{|V_\mathrm{relaxed} - V_\mathrm{original}|}{V_\mathrm{original}}$, where $V_\mathrm{original}$ and $V_\mathrm{relaxed}$ denote the VPA before and after relaxation, respectively. The largest 1\% of volume changes in each crystal system were excluded from the average to remove edge cases associated with failed MACE-MPA-0 relaxations. Smaller VPA changes in high-symmetry systems indicate weaker relaxation-induced structural rearrangements. This trend is consistent with the higher chemical similarity of substitutions shown in (a).}
    \label{fig:volume_change_and_substitution_cost}
\end{figure}

\begin{figure}
    \centering
    \includegraphics[width=\linewidth]{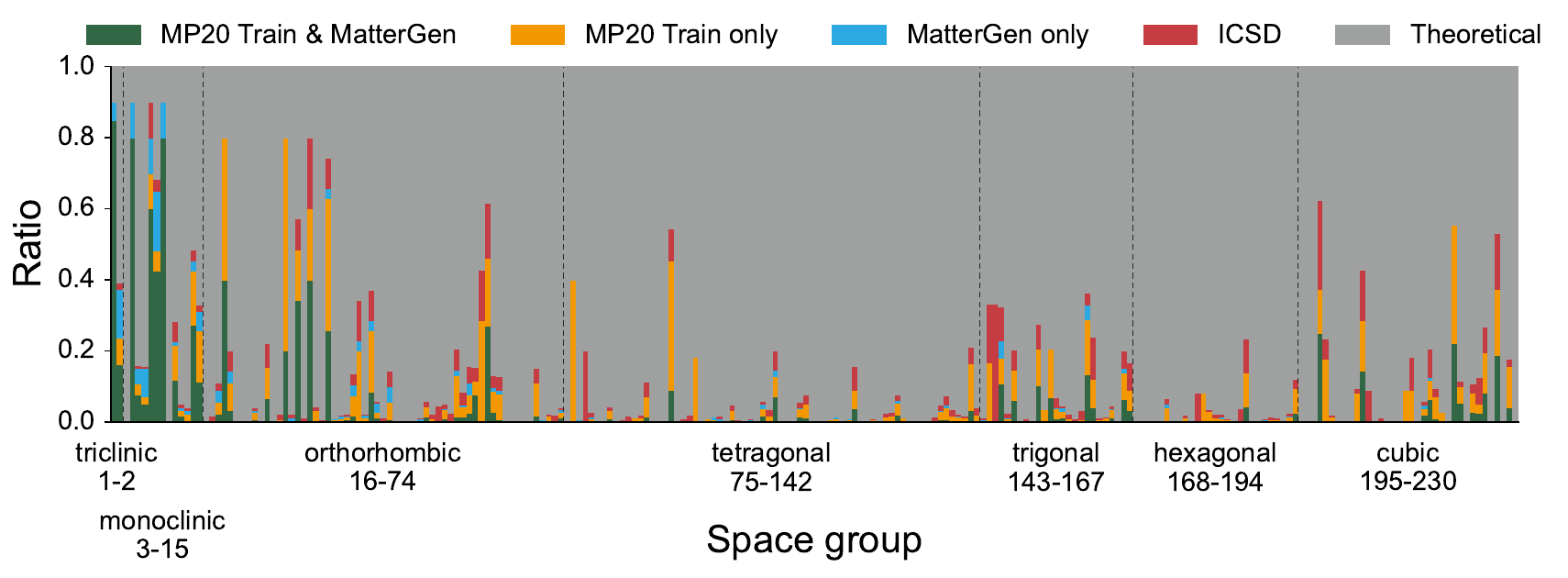}
    \caption{Structural prototype coverage of MP20 training crystals and MatterGen-generated crystals within the full prototype space. For each space group, all symmetry-allowed Wyckoff-label multisets with up to 20 atoms in the primitive unit cell are enumerated. Each multiset is then classified by its occurrence in the MP20 training set or in the MatterGen-generated set. Multisets absent from both sets are further labelled according to their presence in the ICSD database \citep{zagorac2019recent}. The number of MatterGen-generated structures is matched to the training-set size for fair comparison. MP20 covers only a small fraction of the full prototype space, even in high-symmetry regions, leaving many ICSD-observed and unobserved Wyckoff multisets unexplored. Thus, the lower Unmatched rate observed for high-symmetry systems in Figure~\ref{fig:classification_by_crystal_system} is not explained by a scarcity of possible structural prototypes. MatterGen explores prototypes absent from the MP20 training data more effectively in low-symmetry systems, but its exploration is more limited in high-symmetry systems.}
    \label{fig:wyckoff_coverage}
\end{figure} 

Figure~\ref{fig:classification_by_crystal_system} shows the category ratios of MatterGen-generated and MP20 test samples by crystal system, providing further insight into the crystal types represented in each category. 
The Substituted class is subdivided according to the structural similarity criterion used to identify the training structures that ultimately matched each generated or test sample after elemental substitution and relaxation. 
Lattice \& site-based matching provides better initial candidates for low-symmetry systems, whereas Wyckoff-symmetry-based matching is more effective for high-symmetry systems such as hexagonal crystals. 
These results justify the use of complementary structural similarity criteria when selecting training samples similar to a generated structure.

Both MatterGen and the MP20 test set show a decreasing Unmatched ratio with increasing crystal symmetry, although this trend is more pronounced for MatterGen, which will be discussed in the next section. 
This behaviour can be explained by the degree to which symmetry constrains the local coordination environment around each atomic site. 
In high-symmetry systems, atomic sites are subject to stronger symmetry-imposed restrictions on coordination geometry. 
Thus, atoms in generated or test structures are more likely to be chemically similar to their counterparts in matched training structures.
These chemically similar substitutions induce only modest structural changes during relaxation, making the relaxed substituted structures more likely to remain matched to the AI-generated or test structures.
This interpretation is supported by Figure~\ref{fig:volume_change_and_substitution_cost}. 
Figure~\ref{fig:volume_change_and_substitution_cost}(a) shows that substitution costs are lower in high-symmetry systems, meaning that the substituted elements are, on average, more chemically similar to the original atoms.
The higher substitution cost for MatterGen is consistent with the higher compositional novelty of AI-generated samples \citep{negishi2026continuoussunstableunique}.
Figure~\ref{fig:volume_change_and_substitution_cost}(b) reports the average fractional change in volume per atom (VPA), $\frac{|V_\mathrm{relaxed} - V_\mathrm{original}|}{V_\mathrm{original}}$, where $V_\mathrm{original}$ and $V_\mathrm{relaxed}$ denote the VPA before and after relaxation, respectively. 
High-symmetry systems exhibit smaller VPA changes, indicating more modest relaxation-induced structural rearrangements. 
Taken together, these results suggest that the lower Unmatched rate for highly symmetric structures arises from their more strongly constrained local coordination environments, which favor chemically compatible substitutions and help preserve structural similarity during relaxation.

The increase in the Unmatched rate towards lower-symmetry systems is consistent across all generative models tested (SI Figure~\ref*{si:fig:classification_by_crystal_system}). 
Thus, models that generate more low-symmetry crystals tend to exhibit higher overall Unmatched rates. 
This accounts for the high aggregate Unmatched rate of MatterGen in Table~\ref{tab:classification}, as MatterGen generates a larger fraction of low-symmetry crystals. 
Conversely, DiffCSP++ and WyckoffTransformer, which condition generation on space group to explore high-symmetry regions, show lower aggregate Unmatched rates.
When compared within individual crystal systems, however, their Unmatched ratios are comparable to or greater than those of MatterGen (SI Figure~\ref*{si:fig:classification_by_crystal_system}).

The preceding analysis attributed the lower Unmatched ratio in high-symmetry systems to symmetry-constrained coordination environments that favor substitutions between chemically similar elements.
Then, do their fewer degrees of freedom in atomic coordinates imply that the high-symmetry prototype space is inherently limited? 
If this space is already largely covered by the training set, AI-generated or test samples would naturally be classified as Substituted, simply because few novel prototypes remain available.
To test this possibility, we define a structural prototype as a multiset of Wyckoff labels.
For example, cubic perovskite \ce{CsPbI3}, with Pb at $(0,0,0)$, Cs at $(\frac{1}{2},\frac{1}{2},\frac{1}{2})$, and I at $(\frac{1}{2},0,0)$, $(0,\frac{1}{2},0)$, and $(0,0,\frac{1}{2})$, is represented by the multiset $\{\!\!\{1a,1b,3d\}\!\!\}$. 
Figure~\ref{fig:wyckoff_coverage} enumerates, for each space group, all symmetry-allowed Wyckoff-multiset prototypes containing up to 20 atoms in the primitive unit cell and classifies them by their occurrence in the MP20 training set or the MatterGen-generated set. 
Prototypes absent from both sets are further labelled according to whether they appear in the ICSD database.
MP20 covers only a small fraction of the entire prototypes space, including in high-symmetry regions.
Among 2,984,146 symmetry-allowed Wyckoff-label combinations, MP20 contains only 2,598, corresponding to 0.09\%. 
Even for cubic systems, where MatterGen produces no Unmatched crystals, MP20 covers only 4.76\% of the prototype space.
Thus, the MP20 training set does not exhaust the high-symmetry prototype space, indicating that the lower Unmatched rate in Figure~\ref{fig:classification_by_crystal_system} is not due to a lack of possible structural prototypes.
MatterGen seems to be more successful in generating novel prototypes in low-symmetry regions, consistent with the higher Unmatched rate.
Detailed counts of prototypes across the five categories and seven crystal systems are provided in Table~\ref*{si:tab:wyckoff_coverage} in SI.

\subsection{Interpolation and Memorisation in Structural Space}

The preceding discussion examined the decrease in the Unmatched ratio with increasing crystal symmetry, a pattern shared by AI-generated crystals and MP20 test samples. 
However, Figure~\ref{fig:classification_by_crystal_system} also reveals discrepancies between the two sample sets.
First, the Unmatched ratio is even higher for MatterGen-generated crystals than for MP20 test crystals in low-symmetry systems. 
Second, the Duplicate ratio of MatterGen samples increases with crystal symmetry, whereas no such trend is observed for the MP20 test set.
Similar patterns are observed across all tested generative models, although their strength varies (SI Figure~\ref*{si:fig:classification_by_crystal_system}). 
These discrepancies suggest that current generative models do not perfectly learn the underlying distribution represented by MP20. 
This section examines the possible origins of these deviations.

To analyse how AI-generated samples and training samples are distributed in structural space, we first embed them into the space of Average Minimum Distance (AMD) vectors \citep{widdowson2022average}.
An AMD vector is a structural fingerprint for a crystal, where AMD$[k]$ denotes the mean distance from each atom to its $k^\mathrm{th}$ nearest neighbour, averaged over all atoms in the unit cell.
The AMD fingerprint is independent of the choice of unit cell and varies continuously with atomic positions, thereby providing a smooth structural representation. 
It is also sufficiently expressive to distinguish all 229K crystals in the Cambridge Structural Database \citep{widdowson2022average}.

\begin{figure}
    \centering
    \includegraphics[width=\linewidth]{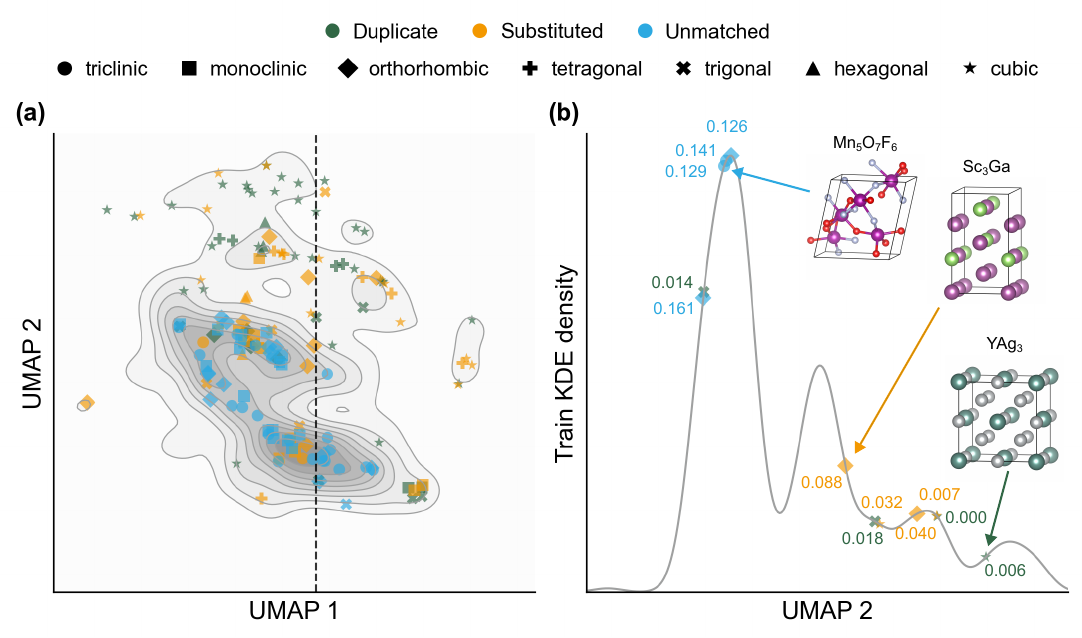}
    \caption{\textbf{a.} UMAP projection of MatterGen-generated structures in Average Minimum Distance (AMD) vector space \citep{widdowson2022average}. For readability, only 50 randomly selected samples from each category are shown. Contours represent the MP20 training-data distribution estimated by kernel density estimation. Unmatched samples cluster in training-dense regions, Duplicate samples occur mainly in training-sparse regions, and Substituted samples appear in both. \textbf{b.} Kernel-density cross section along the dashed line in (a), with samples within $\pm 0.5$ in UMAP 1 also plotted. Numerical labels denote the AMD distance between each generated sample and its nearest training sample. The larger AMD distances in training-dense regions may suggest structural interpolation, leading to Unmatched samples, whereas the shorter distances in training-sparse regions can indicate memorisation, yielding Duplicate samples.}
    \label{fig:umap_kde}
\end{figure}

\begin{figure}
    \centering
    \includegraphics[width=\linewidth]{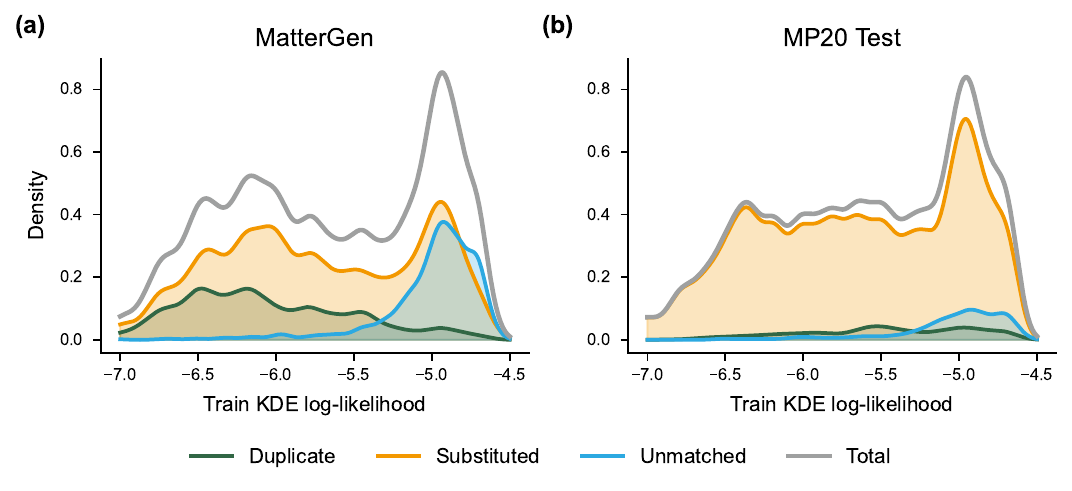}
    \caption{Log-likelihood distributions of MatterGen samples (\textbf{a}) and MP20 test samples (\textbf{b}) by category. The Total curve denotes the aggregate distribution across all categories. Higher log-likelihood values indicate denser regions of the training distribution. MatterGen generates more Unmatched crystals in training-dense regions and more Duplicate crystals in training-sparse regions.}
    \label{fig:likelihood_density}
\end{figure}

Figure~\ref{fig:umap_kde}(a) shows a UMAP projection of 150 randomly selected MatterGen samples (50 from each category), overlaid with contours of the MP20 training-data distribution. 
The UMAP embedding is fitted on the training samples and subsequently used to project the generated samples.
The scatter plot shows that low-symmetry crystals tend to lie in training-dense regions, whereas high-symmetry crystals are more often located in training-sparse regions. 
In other words, low-symmetry crystals generally have a larger number of nearby training samples than high-symmetry crystals.
At first glance, this observation may appear to suggest that low-symmetry crystals, being located in training-dense regions, should be more likely to be classified as Duplicate or Substituted.
However, the opposite turned out to be true.
The scatter plot in Figure~\ref{fig:umap_kde}(a) and its cross section in Figure~\ref{fig:umap_kde}(b) show that Unmatched samples are concentrated in training-dense regions.
By contrast, most Duplicate samples occur in training-sparse regions.
Substituted samples are distributed across both regions.
This pattern is consistent across UMAP random seeds (SI Figure~\ref*{si:fig:umap_umap_seed}) and 150-sample subset-selection seeds (SI Figure~\ref*{si:fig:umap_sample_seed}), and is also observed in the other models examined (SI Figure~\ref*{si:fig:umap_model}).
It can be interpreted as reflecting interpolation in dense regions of the training distribution and memorisation in sparse regions.
In training-dense regions, the models seem to learn local structural environments from abundant training examples and use this knowledge to generate Unmatched crystals that interpolate between training samples.
The distance annotations in Figure~\ref{fig:umap_kde}(b) support this interpolation view, showing that Unmatched samples tend to be farther from their nearest training structures despite lying in training-dense regions.
Moreover, this interpolation behaviour is more pronounced for MatterGen samples than for MP20 test samples, as indicated by the higher Unmatched peak for MatterGen in Figure~\ref{fig:likelihood_density}.
This helps explain the higher Unmatched rate of MatterGen in low-symmetry systems observed in Figure~\ref{fig:classification_by_crystal_system}.
Conversely, in training-sparse regions, generated samples tend to lie closer to their nearest training structures, as shown by the distance annotations in Figure~\ref{fig:umap_kde}(b).
Interpolation appears less common, while copying or memorisation of training structures occurs more frequently.
This memorisation in training-sparse regions is also a behaviour specific to MatterGen, as shown in Figure~\ref{fig:likelihood_density}, and accounts for the higher Duplicate ratio in higher-symmetry systems in Figure~\ref{fig:classification_by_crystal_system}.
Overall, the learned model distribution seems to be smooth and interpolative in training-dense regions, but sharp and more prone to memorisation in training-sparse regions.

The patterns observed in Figures~\ref{fig:umap_kde} and \ref{fig:likelihood_density} are not unique to MatterGen, but recur across the tested generative models (SI Figures~\ref*{si:fig:umap_model} and \ref*{si:fig:likelihood_density}). 
Similar interpolation--memorisation behaviour has been analysed theoretically for generative models, especially diffusion models, through the lens of distributional smoothness and sharpness \citep{scarvelis2025closedformdiffusionmodels,jeon2025understanding,chen2026on}. 
Our findings indicate that this phenomenon extends to crystal generation.

\subsection{Examples of Unmatched Crystals}

\begin{figure}
    \centering
    \includegraphics[width=\linewidth]{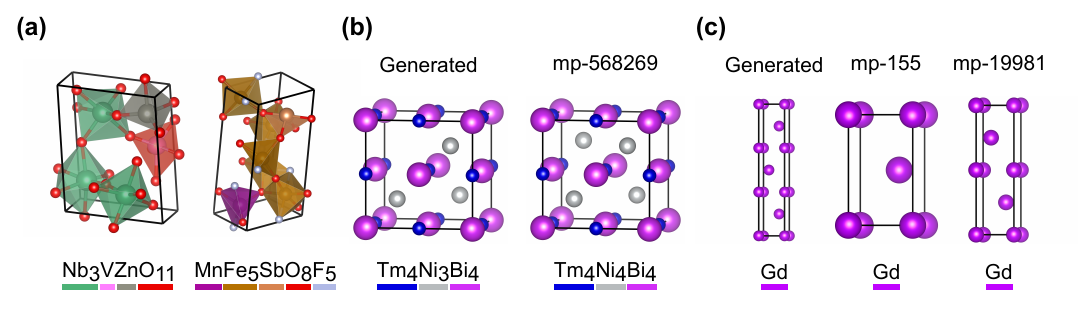}
    \caption{\textbf{a.} Representative MatterGen-generated crystals classified as Unmatched. Locally plausible coordination polyhedra are connected by O and F atoms, yielding complex global structures. \textbf{b, c.} Unmatched MatterGen crystals that appear to arise from non-substitutional modifications of training structures: removal of one Ni atom from a training sample (\textbf{b}) and stacking of two training samples (\textbf{c}). Although classified as Unmatched in the current workflow, such structures may be accessible via classical genetic-algorithm-like structure-search methods.}
    \label{fig:unmatched_examples}
\end{figure}

The analyses presented so far were based mainly on aggregate statistics. 
Here, we examine representative Unmatched crystals to better understand the structures captured by this category. Figure~\ref{fig:unmatched_examples}(a) shows two MatterGen-generated Unmatched crystals containing O or F, which are enriched in the Unmatched class. 
Specifically, O appears in 39.7\% of MatterGen Unmatched samples, compared with 13.5\% of Substituted or Duplicate samples and 25.9\% of the MP20 training set. 
Similarly, F appears in 15.0\% of Unmatched samples, compared with 4.6\% of Substituted or Duplicate samples and 6.9\% of the training set.
The visualised crystals have complex low-symmetry structures that are difficult to match to training samples. 
Nevertheless, their local atomic environments appear chemically reasonable, although slight structural distortions are present.
In the left structure, Nb, V, and Zn form octahedral, square-pyramidal, and tetrahedral environments, respectively. 
In the right structure, Mn is square-pyramidal, Fe shows mixed octahedral, tetrahedral, and square-pyramidal coordination, and Sb adopts a seesaw-like geometry, likely reflecting its lone pair. 
These structures can be interpreted as examples of interpolation in training-dense regions. Because the training data contain many oxides and fluorides, the models can learn typical metal coordination environments and the diverse ways in which coordination polyhedra are connected, including corner-, edge-, and face-sharing motifs. 
The models then appear to use this learned structural knowledge to generate locally plausible but globally novel complex structures that are classified as Unmatched.

A closer inspection of Unmatched samples also highlights cases that fall outside the scope of our current classification workflow. 
Figure~\ref{fig:unmatched_examples}(b) shows an Unmatched crystal that is obtainable from a similar half-Heusler training structure by removing a single Ni atom. 
Likewise, the Unmatched crystal in Figure~\ref{fig:unmatched_examples}(c) is formed by stacking two different training structures. 
Since our workflow is designed to detect only substitution-derived structures, these samples are classified as Unmatched. 
Nevertheless, they may be easily reproducible by classical evolutionary algorithms for structure search, which often include atom insertion/deletion operations, such as \texttt{NumAtomsMut} in GASP \citep{Tipton_2013} and \texttt{permutomic} in XtalOpt \citep{LONIE2011372, HAJINAZAR2026109910}, as well as structural crossover operations, such as \texttt{heredity} in USPEX \citep{GLASS2006713} and \texttt{crossover} in XtalOpt.
Identifying such non-substitutional relationships is therefore a promising direction for future work.

\section{Discussion}

We have introduced a workflow to assess whether generative AI models explore materials space beyond simple elemental substitution in known crystal structures. 
The workflow first identifies AI-generated crystals that exactly match training structures as duplicates. 
For the remaining samples, it selects structurally and chemically similar training structures, reducing the cost of exhaustive comparison while favouring chemically plausible substitutions. 
These selected training samples are then subjected to elemental substitution and relaxation, and the resulting structures are compared with the AI-generated crystal to determine whether it is substitution-derived or not.

Applying the workflow to chemically valid and metastable crystals sampled from representative generative models revealed that 81--92\% are either training duplicates or substitution-derived structures.
This trend is especially pronounced in high-symmetry crystal systems, suggesting that current crystal generative models show limited expansion beyond substitution-based search in these regions.
The high substitution-derived ratio can be attributed to symmetry-constrained local coordination environments, which promote substitutions between chemically similar elements.
However, the fewer degrees of freedom in atomic coordinates do not imply that the high-symmetry prototype space is inherently limited: many structural prototypes, including both experimentally observed and theoretical prototypes, are absent from the MP20 training data. 
Nevertheless, the capability of these models to explore novel prototypes remains limited in high-symmetry systems, highlighting a current limitation of crystal generative models.

Analysis in the structural space, using an AMD descriptor, provides an interpretation of our classification results from the perspective of interpolation and memorisation.
High-symmetry crystals tend to lie in training-sparse regions, where the learned distribution appears sharp and memorisation-prone, leading to higher duplicate rates. 
Low-symmetry crystals, in contrast, cluster in training-dense regions, where models learn a smoother, more interpolative distribution. 
This enables them to combine local structural environments learned from abundant training samples into crystals with plausible local structures and complex global architectures that are novel.
Oxides and fluorides, in which O and F link coordination polyhedra through diverse connectivity patterns, provide examples of such low-symmetry structures.

A limitation of this study is that it only considers how large the search space of generative models is compared to substitution-based algorithms.
A missing perspective is how efficiently algorithms can explore their search space, an aspect in which generative models may have a clear advantage.
Once trained on databases of known crystals, generative models can rapidly generate large numbers of candidate compounds through inference, which is generally faster than conventional evolutionary algorithms \citep{cheng2025ai}.
They may therefore provide a more efficient way to sample the high-symmetry search space already accessible through substitution-based strategies.
However, our results suggest that they do not fundamentally expand the boundaries of this search space in high-symmetry regions.
One limitation of our study is that the present workflow detects only substitution-derived crystals. 
It does not account for other operations that are commonly used in conventional evolutionary algorithms for crystals, such as atom insertion or deletion and crossover between different structures. 
Incorporating these operations, with an appropriate cost function, would enable a deeper comparison between the search spaces accessible to generative models and evolutionary techniques.

\section{Methods}
\label{sec:method}

As illustrated in Figure~\ref{fig:workflow}, our workflow consists of five steps: duplication check, structural filtering, chemical filtering, substitution with relaxation, and substitution check. 
In the duplication-check and substitution-check stages, a generated crystal is compared with the training structures and the substituted structures, respectively. 
These comparisons are performed using \texttt{StructureMatcher} in \texttt{pymatgen} with its default threshold values. 
Figure~\ref*{si:fig:sm_sensitivity} and Table~\ref*{si:tab:sm_sensitivity} in SI show how the classification results vary with the threshold parameters.
The overall classification trends remain consistent across the tested threshold ranges, including thresholds up to four times stricter than the default values.
In the substitution stage, structural relaxation is performed using the MACE-MPA-0 force field \citep{batatia2025foundation}. 
We use the Limited-memory BFGS algorithm \citep{liu1989limited} with a force convergence threshold of $10^{-3}$~[eV/\AA] and a maximum of 1000 steps. The following sections describe the structural- and chemical-filtering steps in detail.

\subsection{Structural Similarity Criteria}

In our workflow, if a given AI-generated crystal $G$ is not a duplicate of any training crystal, the next step is to select training crystals that are structurally similar to $G$ (Figure~\ref{fig:workflow}). 
This filtering step is necessary because optimal atomic-site mappings between $G$ and all training samples are unavailable a priori, making it difficult to apply substitutions directly to every pair. 
It also reduces the number of structural relaxations required in later stages, thereby lowering the computational cost.
Our workflow adopts two different criteria of structural similarity: difference in cell shape and fractional coordinates, and matching of space group and Wyckoff letters.

The first criterion selects a training structure $T$ as similar to $G$ when the two structures can be represented by similarly shaped unit cells and closely matching fractional atomic coordinates. 
Specifically, the atomic species in $T$ and $G$ are first anonymised, and the two structures are then compared using \texttt{StructureMatcher} in \texttt{pymatgen}. 
This procedure systematically searches over possible lattice transformations and translations to align one structure with the other.
A match is confirmed if an alignment satisfies the following conditions: the fractional difference in lattice-vector lengths is below \texttt{ltol}, the difference in lattice-vector angles is below \texttt{angle\_tol} [degrees], and the fractional coordinate difference is below \texttt{stol}$ * \sqrt[3]{V/N}$, where $V$ and $N$ denote the cell volume and number of atoms, respectively. 
Unless otherwise stated, we use the package default values, \texttt{ltol=0.2}, \texttt{stol=0.3}, and \texttt{angle\_tol=5}.

The second criterion uses space-group and Wyckoff-label matching. 
We first briefly introduce these concepts.
A space group is the set of symmetry operations, including translations and rotations, that leave a crystal invariant.
Every crystal is assigned to one of 230 space groups. 
Within a given space group, a Wyckoff position specifies a set of symmetrically equivalent sites: placing an atom at one representative site generates all symmetry-equivalent atoms through the space-group operations.
Each Wyckoff position is characterised by a Wyckoff label, which consists of its multiplicity and a letter denoting its site symmetry.
Therefore, the space group together with the occupied Wyckoff labels provides a compact representation of the crystal’s symmetry and atomic arrangement. 
For example, cubic perovskite \ce{CsPbI3} belongs to space group 221 and is represented by the Wyckoff-label multiset $\{\!\!\{1a,1b,3d\}\!\!\}$. 
Our second structural-similarity criterion identifies a training structure $T$ as similar to $G$ if $T$ and $G$ belong to the same space group and have identical multisets of Wyckoff labels.
It is worth noting that the Wyckoff-label multiset is not always unique for a given crystal structure. 
Even after the lattice vectors are fixed by Niggli reduction, the choice of origin can change the assigned Wyckoff labels. 
For example, \ce{CsPbI3}, which is represented as $\{\!\!\{1a,1b,3d\}\!\!\}$ in one origin choice, can equivalently be represented as $\{\!\!\{1a,1b,3c\}\!\!\}$ after shifting the origin by $(0.5,0.5,0.5)$. 
Our Wyckoff-label matching pipeline therefore uses the Euclidean normaliser of the space group to match label multisets that are equivalent up to such origin shifts.

These two criteria appear to operate complementarily.
As shown in Figure~\ref{fig:classification_by_crystal_system}, lattice-and-site matching identifies better initial candidates for low-symmetry systems, whereas Wyckoff-based matching is more effective for high-symmetry systems. 
This is because low-symmetry structures retain greater degrees of freedom in lattice parameters and atomic coordinates, so crystals sharing the same space group and Wyckoff-label multiset may still differ substantially. 
In high-symmetry systems, by contrast, the space group and Wyckoff labels impose stronger constraints and therefore specify much of the crystal structure.

\subsection{Substitution Cost Based on Modified Pettifor Chemical Scale}

Once structurally similar training crystals are selected for a non-duplicate generated structure $G$, our workflow further filters them based on the chemical similarity between atoms.
This step reduces the number of post-substitution relaxations, since some generated crystals can match hundreds of training structures; for instance, a rocksalt-shape $G$ would match all rocksalt-shape training samples under the Wyckoff-based criterion. 
Chemical filtering also discourages substitutions between dissimilar elements, which can cause large relaxation-induced structural changes and may be avoided in conventional substitution-based search algorithms.

The chemical similarity between $G$ and a structurally similar training structure $T$ is defined as the average difference in the modified Pettifor chemical scale \citep{glawe2016optimal} over matched atom pairs.
The modified Pettifor chemical scale provides a 1D ordering of elements up to atomic number 103 that reflects chemical similarity.
For example, subsequences such as ``\ce{Au}, \ce{Ag}, \ce{Cu}, \ce{Ni}, \ce{Co}, \ce{Fe}" and ``\ce{O}, \ce{At}, \ce{I}, \ce{Br}, \ce{Cl}, \ce{F}" appear in this ordering.
The sequence is optimised so that elements frequently interchangeable in the ICSD database are positioned close to one another, making it suitable for quantifying the chemical plausibility of elemental
substitutions.
Using this scale, the substitution cost between elements $x$ and $y$ is defined as $c(x, y) = |s(x) - s(y)| \in [0, 102]$, where $s(\cdot)$ denotes the element's positions in the sequence.
Using this element-wise cost, the substitution cost between $G$ and $T$ is defined as
\begin{equation*}
    C(G, T) = \frac{1}{N} \sum_{i=1}^{N} c(G(i), T(i)),
\end{equation*}
where $G(i)$ and $T(i)$ are the elements at the $i$-th matched atomic-site pair, and $N$ is the number of atoms in the unit cell.
The mapping between atomic sites in $G$ and $T$ is obtained naturally from the preceding structural-matching step.
In the chemical-filtering step of our workflow, the top-$k$ candidates with the lowest substitution costs from each structural-similarity criterion are selected for the subsequent substitution step. 
Here, $k$ is a hyperparameter, and we use $k=3$ as the default value. 
Larger values of $k$ enable a more rigorous assessment of whether $G$ is substitution-derived, but the computational time required for the subsequent structural relaxations increases linearly with $k$. 
Figure~\ref*{si:fig:topk_sensitivity} presents a sensitivity analysis with respect to $k$.

\section{Data Availability}

The datasets analysed during the current study are available in a HuggingFace repository at \url{https://huggingface.co/datasets/masahiro-negishi/xtaledit}. 

\section{Code Availability}

The underlying code for this study is available in a GitHub repository at \url{https://github.com/WMD-group/xtaledit}.

\newpage

\bibliography{sn-bibliography}%

\paragraph{Acknowledgements}
This work was supported by the AIchemy Hub through EPSRC grants EP/Y028775/1 and EP/Y028759/1, and by an Imperial College President’s PhD Scholarship. We acknowledge the EuroHPC Joint Undertaking for providing access to the EuroHPC supercomputer LEONARDO, hosted by CINECA in Italy and the LEONARDO consortium, through a EuroHPC Extreme Access call. We also thank Katharina Ueltzen from the Federal Institute for Materials Research and Testing for valuable feedback on the experimental design.

\paragraph{Author Contributions}
M.N. and A.W. conceived the study. M.N. curated the data, developed the methodology and software, performed the investigation, formal analysis and validation, visualized the results, and wrote the original draft. A.W. acquired funding, provided resources, administered the project, and supervised the work. Both authors reviewed and edited the manuscript.

\paragraph{Competing Interests}
A.W. is Chief Scientific Officer at CuspAI. M.N. declares no competing interests.

\newpage

\begin{center}
    \vspace*{2cm}
    {\Huge \textbf{Supporting Information}} \\
    \vspace{1cm}
\end{center}

\etocdepthtag.toc{SI}
\etocsettagdepth{mt}{none}
\etocsettagdepth{SI}{subsection}
\etocsettocstyle{}{} 

\setcounter{section}{0}
\setcounter{page}{1}
\setcounter{figure}{0}
\setcounter{table}{0}
\setcounter{equation}{0}
\setcounter{theorem}{0}

\renewcommand{\thesection}{S\arabic{section}}
\renewcommand{\thefigure}{S\arabic{figure}}
\renewcommand{\thetable}{S\arabic{table}}
\renewcommand{\theequation}{S\arabic{equation}}
\renewcommand{\thetheorem}{S\arabic{theorem}}

\renewcommand{\theHsection}{SI.\thesection}
\renewcommand{\theHfigure}{SI.\thefigure}
\renewcommand{\theHtable}{SI.\thetable}
\renewcommand{\theHequation}{SI.\theequation}
\renewcommand{\theHtheorem}{SI.\thetheorem}

\makeatletter
\renewcommand{\citenumfont}[1]{S#1}
\renewcommand{\bibnumfmt}[1]{[S#1]}
\makeatother

Additional figures and tables for ``Substitution-Based Analysis of Structural Novelty for Generative Materials Models".

\begin{figure}[h]
    \centering
    \includegraphics[width=\linewidth]{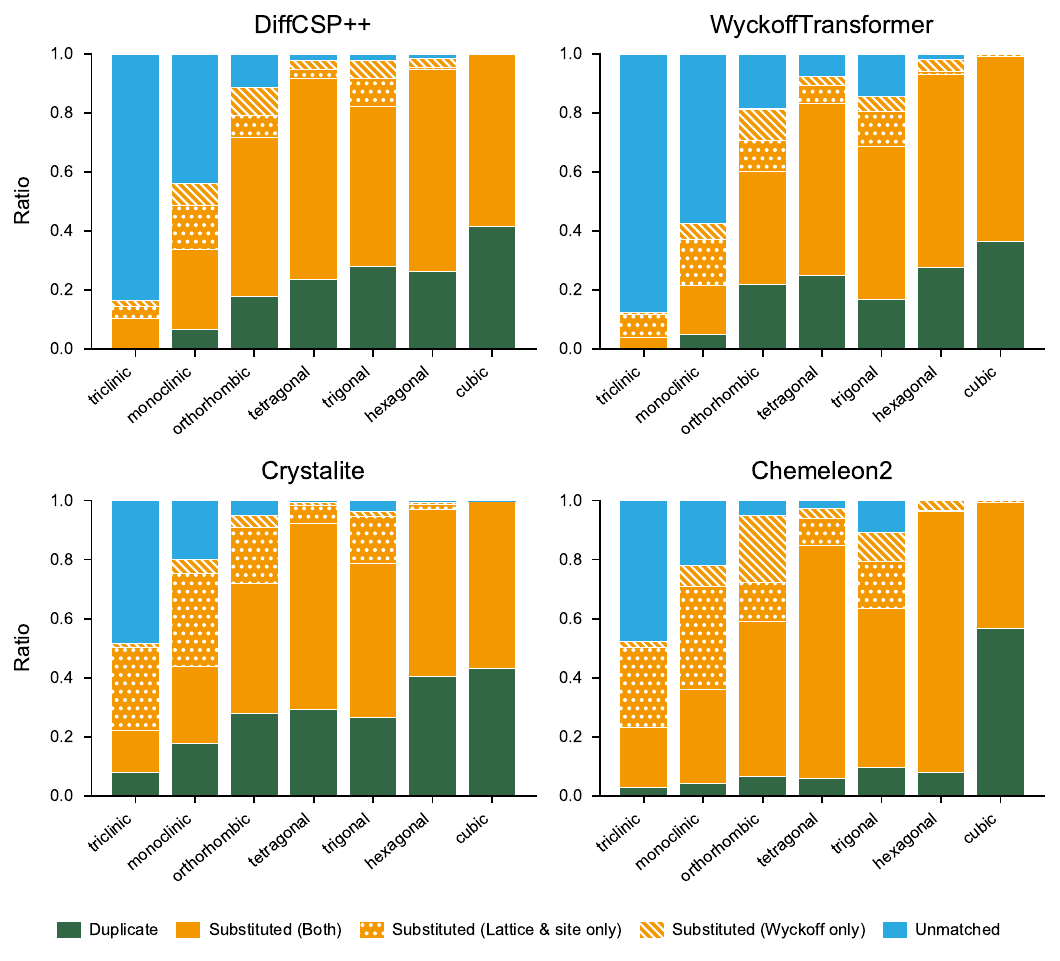}
    \caption{Classification of samples from four generative models by crystal system. Parentheses in the legend indicate the structural similarity criterion used to identify the training structures that ultimately matched each generated sample after elemental substitution and relaxation. All models show higher Unmatched rates in lower-symmetry systems and higher Duplicate rates in higher-symmetry systems, although the magnitude of these trends differs. Chemeleon2 exhibits relatively low Duplicate rates in all non-cubic systems, consistent with its reinforcement-learning fine-tuning designed to penalise duplicates.}
    \label{si:fig:classification_by_crystal_system}
\end{figure}

\begin{table}[h]
    \caption{Structural prototype coverage of MP20 training and MatterGen-generated crystals within the full prototype space. For each crystal system, we enumerate all symmetry-allowed Wyckoff-label multisets containing up to 20 atoms in the primitive unit cell and classify each by its occurrence in the MP20 training set, the MatterGen-generated set, or neither. Multisets absent from both sets are further labelled by their presence in the ICSD database. The number of MatterGen-generated structures is matched to the training-set size for fair comparison.}
    \label{si:tab:wyckoff_coverage}
    \centering
    \begin{tabular}{@{}lcccccc@{}}
        \toprule
        Crystal system & \makecell{MP20 Train\\\& MatterGen} &
        \makecell{MP20 Train\\only} & \makecell{MatterGen\\only} & ICSD & Theoretical & Total \\
        \midrule
        triclinic & 83 & 30 & 57 & 7 & 248 & 425 \\
        monoclinic & 311 & 216 & 257 & 107 & 144,558 & 145,449 \\
        orthorhombic & 252 & 529 & 388 & 333 & 2,642,443 & 2,643,945 \\
        tetragonal & 178 & 362 & 109 & 206 & 137,908 & 138,763 \\
        trigonal & 129 & 192 & 43 & 129 & 16,934 & 17,427 \\
        hexagonal & 49 & 147 & 20 & 96 & 35,306 & 35,618 \\
        cubic & 46 & 74 & 3 & 43 & 2353 & 2,519 \\
        \bottomrule
    \end{tabular}
\end{table}

\begin{figure}[h]
    \centering
    \includegraphics[width=\linewidth]{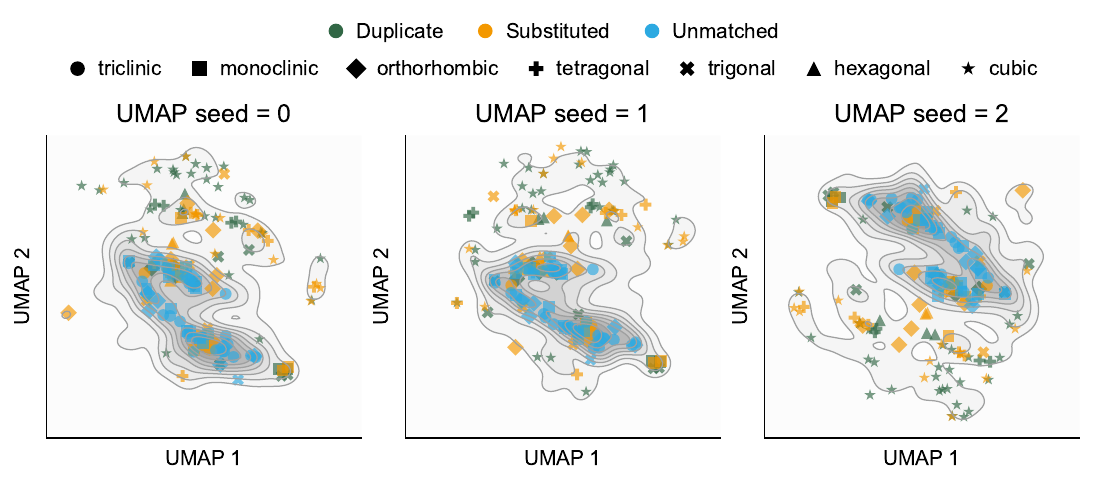}
    \caption{UMAP projections of MatterGen-generated structures in Average Minimum Distance (AMD) vector space obtained with different UMAP random seeds. For readability, only 50 randomly selected samples from each category are shown. Contours represent the MP20 training-data distribution estimated by kernel density estimation. Across all seeds, low-symmetry and Unmatched samples consistently cluster in training-dense regions, while high-symmetry and Duplicate samples occur mainly in training-sparse regions.}
    \label{si:fig:umap_umap_seed}
\end{figure}

\begin{figure}[h]
    \centering
    \includegraphics[width=\linewidth]{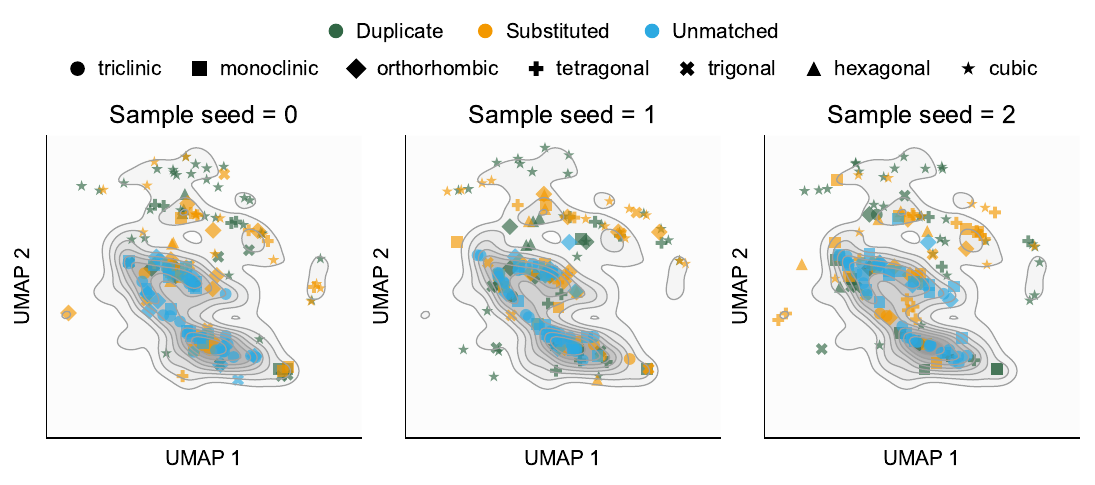}
    \caption{UMAP projections of different 150-sample subsets of MatterGen-generated structures in Average Minimum Distance (AMD) vector space. Contours indicate the MP20 training-data distribution estimated by kernel density estimation. Regardless of the random seed used for subset selection, low-symmetry and Unmatched samples consistently cluster in training-dense regions, while high-symmetry and Duplicate samples are predominantly found in training-sparse regions.}
    \label{si:fig:umap_sample_seed}
\end{figure}

\begin{figure}[h]
    \centering
    \includegraphics[width=\linewidth]{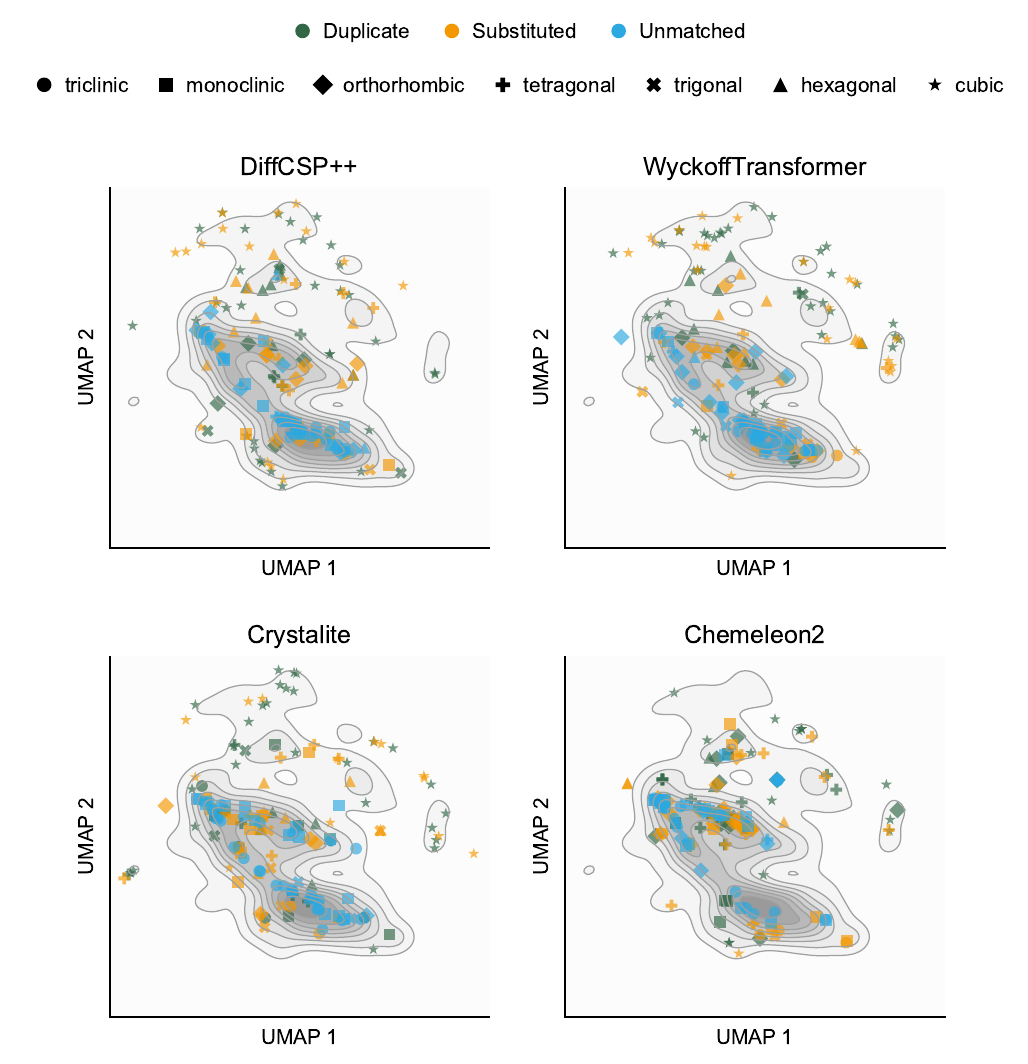}
    \caption{UMAP projections of samples from four generative models in Average Minimum Distance (AMD) vector space. For clarity, only 50 randomly selected samples from each category are displayed. Contours indicate the MP20 training-data distribution estimated by kernel density estimation. Across all models, low-symmetry and Unmatched samples consistently cluster in training-dense regions, while high-symmetry and Duplicate samples are found mainly in training-sparse regions. Chemeleon2 samples appear more spatially concentrated than those of the other models, likely reflecting its reinforcement-learning fine-tuning.}
    \label{si:fig:umap_model}
\end{figure}

\begin{figure}[h]
    \centering
    \includegraphics[width=\linewidth]{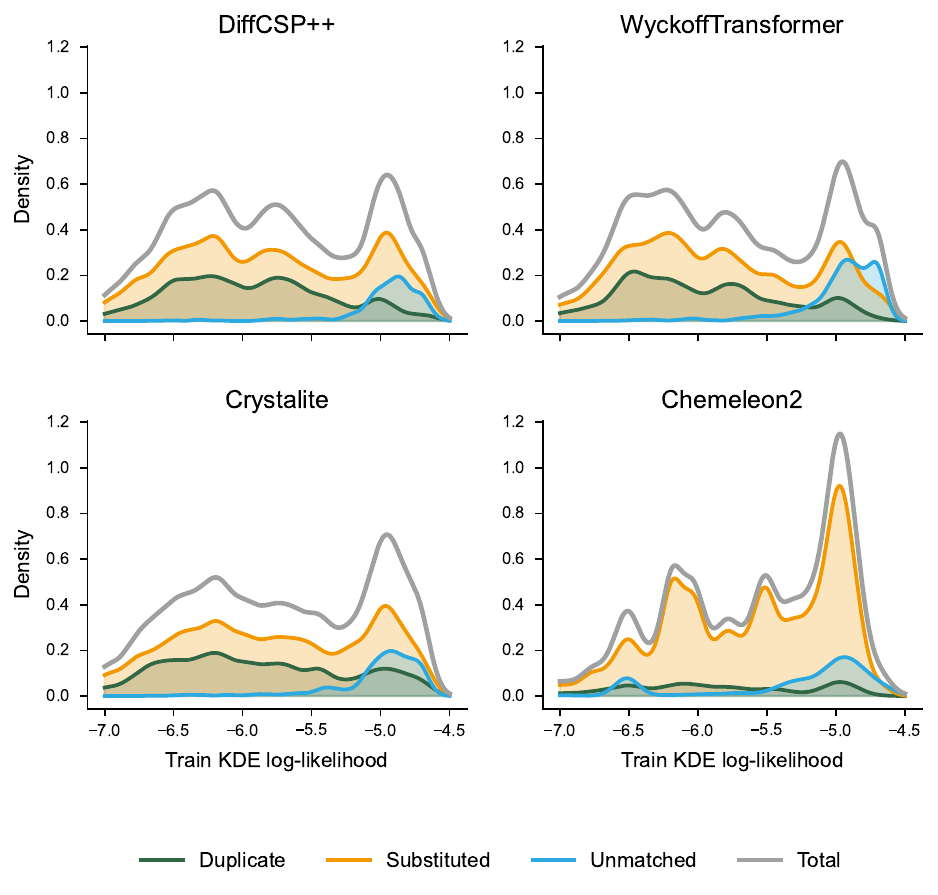}
    \caption{Log-likelihood distributions of samples from different generative models. The Total curve represents the aggregate distribution across categories, and higher log-likelihood values indicate training-dense regions. DiffCSP++, WyckoffTransformer, and Crystalite show distributions similar to MatterGen (Figure~\ref*{fig:likelihood_density}(a) in the main text), with Unmatched samples peaking in training-dense regions and Duplicate samples concentrated in training-sparse regions. Chemeleon2 shows lower Duplicate density, consistent with reinforcement-learning fine-tuning that penalises duplicates. Its Unmatched distribution peaks in training-dense regions, as observed for the other models, but it also exhibits a small additional Unmatched peak in training-sparse regions that is absent from the other models.}
    \label{si:fig:likelihood_density}
\end{figure}

\begin{table}[h]
    \caption{Classification results for MatterGen-generated crystals under increasingly stringent \texttt{pymatgen} \texttt{StructureMatcher} thresholds. As expected, stricter thresholds slightly increase the Unmatched rate and decrease the Duplicate rate, but the overall changes are small.}
    \label{si:tab:sm_sensitivity}
    \centering
    \begin{tabular}{@{}lccccccc@{}}
        \toprule
        & \texttt{ltol} & \texttt{stol} & \texttt{angle\_tol} & Duplicate (\%) & Substituted (\%) & Unmatched (\%) \\
        \midrule
        Default & 0.2 & 0.3 & 5 & 20.5 & 60.6 & 18.9 \\
        3/4 * Default & 0.15 & 0.225 & 3.75 & 19.8 & 59.8 & 20.4 \\
        1/2 * Default & 0.1 & 0.15 & 2.5 & 19.1 & 59.1 & 21.8 \\
        1/4 * Default & 0.05 & 0.075 & 1.25 & 18.1 & 58.8 & 23.1 \\
        \bottomrule
    \end{tabular}
\end{table}

\begin{figure}[h]
    \centering
    \includegraphics[width=\linewidth]{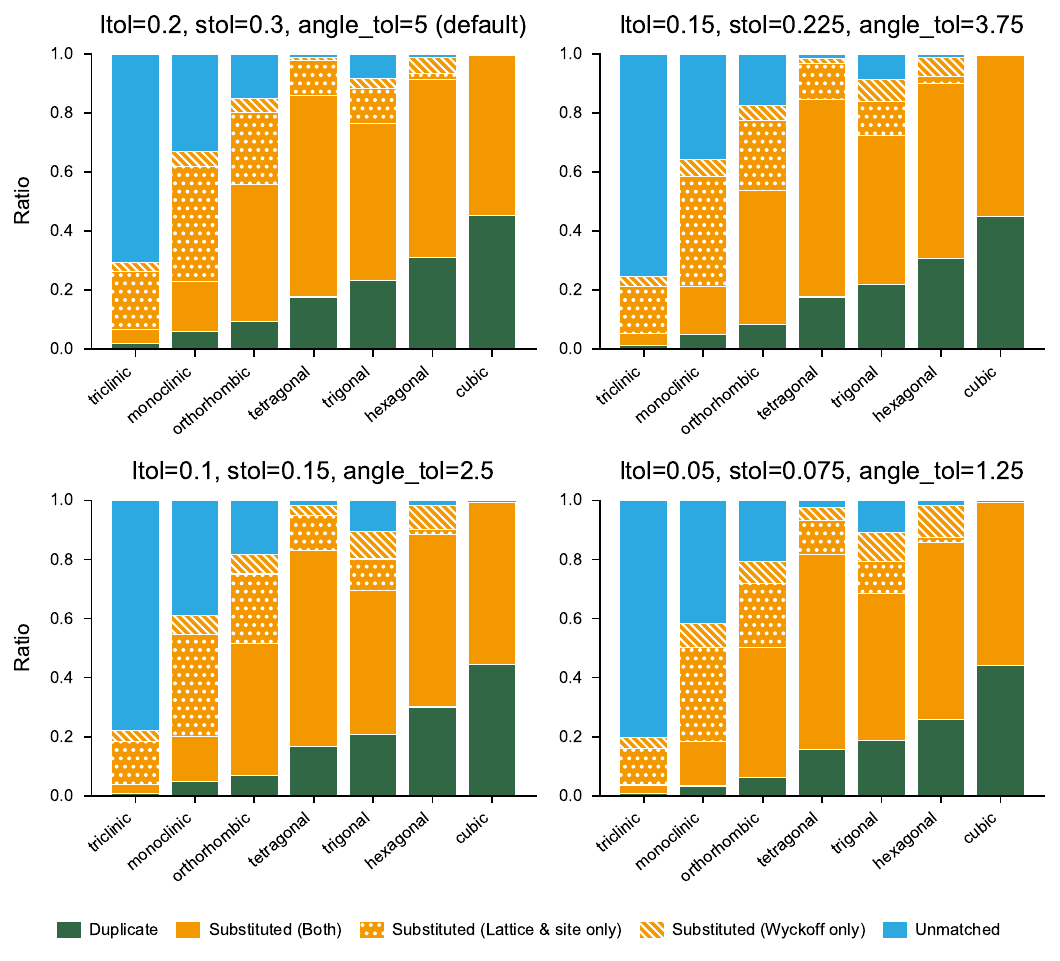}
    \caption{Classification of MatterGen-generated samples by crystal system under different \texttt{pymatgen} \texttt{StructureMatcher} thresholds. The upper-left panel uses the default thresholds for the duplication and substitution checks and is identical to Figure~\ref*{fig:classification_by_crystal_system}(a). The remaining panels use increasingly stringent thresholds. Stricter thresholds slightly increase the Unmatched rate and decrease the Duplicate rate, but the overall classification trends remain robust.}
    \label{si:fig:sm_sensitivity}
\end{figure}

\begin{figure}[h]
    \centering
    \includegraphics[width=\linewidth]{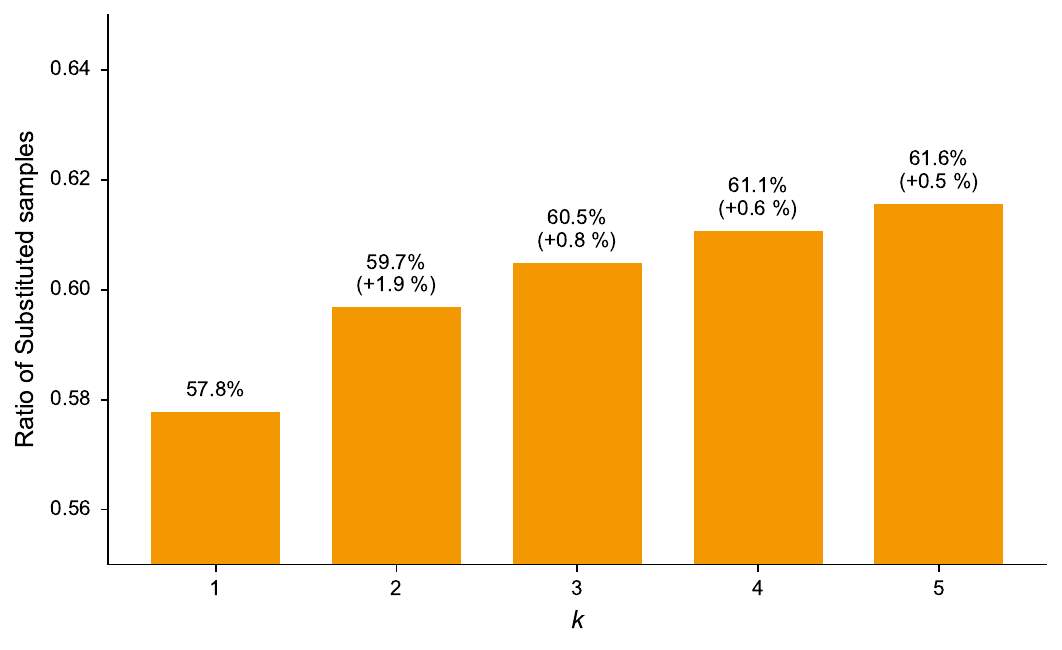}
    \caption{Fraction of MatterGen-generated samples classified as Substituted as a function of $k$, the number of training structures retained from each structural-similarity criterion for substitution and relaxation. Increasing $k$ raises the Substituted ratio by evaluating more candidate training structures, but the improvement saturates while the relaxation cost grows linearly. We therefore adopt $k=3$ by default.}
    \label{si:fig:topk_sensitivity}
\end{figure}

\end{document}